\documentclass[conference]{IEEEtran}
\IEEEoverridecommandlockouts

\usepackage{cite}
\usepackage{amsmath,amssymb,amsfonts}
\usepackage{graphicx}
\usepackage{booktabs}
\usepackage{multirow}
\usepackage{array}
\usepackage{textcomp}
\usepackage{xcolor}
\usepackage{float}
\usepackage{enumitem}
\usepackage[hidelinks]{hyperref}

\graphicspath{{figures/}}

\begin{document}

\title{Evaluating CUDA Tile for AI Workloads on Hopper and Blackwell GPUs}

 \author{ \IEEEauthorblockN{Divakar Kumar Yadav} \IEEEauthorblockA{\textit{University of Wisconsin--Milwaukee} \\ Milwaukee, WI, USA \\ \texttt{dkyadav@uwm.edu}}
 \and \IEEEauthorblockN{Tian Zhao} \IEEEauthorblockA{\textit{University of Wisconsin--Milwaukee} \\ Milwaukee, WI, USA \\ \texttt{tzhao@uwm.edu}} 
 \and \IEEEauthorblockN{Deepak Kumar} \IEEEauthorblockA{\textit{Illinois Institute of Technology} \\ Chicago, IL, USA \\ \texttt{dkumar15@hawk.illinoistech.edu}}
 }

\maketitle

% =============================================================================
% ABSTRACT
% =============================================================================
\begin{abstract}
NVIDIA's CUDA Tile (CuTile) introduces a Python-based, tile-centric abstraction for GPU kernel development that aims to simplify programming while retaining Tensor Core and Tensor Memory Accelerator (TMA) efficiency on modern GPUs. We present the first independent, cross-architecture evaluation of CuTile against established approaches -- cuBLAS, Triton, WMMA, and raw SIMT -- on three NVIDIA GPUs spanning Hopper and Blackwell: H100 NVL, B200, and RTX PRO 6000 Blackwell Server Edition. We benchmark representative AI workloads, including GEMM, fused multi-head attention, and end-to-end LLM inference in BF16/FP16 precision, to assess both performance and portability.

Our results show that CuTile's effectiveness is strongly workload- and architecture-dependent. On datacenter-class Blackwell (B200), CuTile achieves up to 1{,}007~TFLOP/s for fused attention, outperforming FlashAttention-2 by 2.5$\times$ while requiring only 60 lines of Python kernel code. For GEMM, CuTile reaches 52--79\% of cuBLAS performance in 22 lines of code (versus 123 for WMMA), making it a practical replacement for hand-written CUDA kernels but not yet for vendor-optimized libraries. However, the same CuTile attention kernel achieves only 53\% of FlashAttention-2 throughput on RTX PRO 6000 (\texttt{sm\_120}), exposing significant cross-architecture optimization gaps. In contrast, Triton sustains 62--101\% of cuBLAS performance across all tested platforms without architecture-specific tuning, demonstrating substantially stronger portability.
\end{abstract}


% =============================================================================
% 1. INTRODUCTION
% =============================================================================

\section{Introduction}

Writing high-performance GPU kernels remains challenging.
The transformer architecture~\cite{vaswani2017attention} has driven language models to hundreds of billions of parameters~\cite{brown2020gpt3,chowdhery2023palm,llama}, and modern serving systems---including vLLM~\cite{vllm}, TensorRT-LLM~\cite{tensorrt_llm}, and Megatron-LM~\cite{shoeybi2020megatron}---rely on hand-optimized kernels such as FlashAttention-2~\cite{dao2023flashattention2} and CUTLASS~\cite{cutlass} to approach peak hardware utilization.
These kernels typically span hundreds to thousands of lines of architecture-specific CUDA code and often require extensive retuning or redesign with each new GPU generation.

In late 2025, NVIDIA introduced CUDA Tile (CuTile), a Python-based, tile-centric programming model~\cite{nvidia2025cutile} intended to reduce this engineering burden.
CuTile abstracts warps, registers, and shared memory behind high-level primitives (\texttt{ct.load}, \texttt{ct.mma}, \texttt{ct.store}) while still leveraging Tensor Cores and the Tensor Memory Accelerator (TMA) on Blackwell GPUs.
The promise is compelling: express a Tensor Core kernel in tens of lines of Python rather than hundreds of lines of CUDA~C++.
Whether this promise holds in practice, however, remains an open question.

This paper addresses the practical question faced by GPU kernel developers: \emph{Should one switch to CuTile, and if so, for which workloads, on which GPUs, and with what trade-offs?}

We evaluate CuTile head-to-head against established alternatives spanning vendor libraries, high-level DSLs, and hand-written CUDA:
\begin{enumerate}[nosep,leftmargin=*]
  \item \textbf{cuBLAS}~\cite{cublas} --- NVIDIA's closed-source, auto-tuned BLAS library (1~LOC; performance upper bound)
  \item \textbf{Triton}~\cite{tillet2019triton} --- OpenAI's Python-based GPU compiler and leading open-source alternative ($\sim$53--62~LOC)
  \item \textbf{CUDA Tile (CuTile)}~\cite{nvidia2025cutile} --- NVIDIA's tile-centric Python DSL ($\sim$22--60~LOC; \emph{subject of this study})
  \item \textbf{WMMA}~\cite{nvidia_wmma} --- Hand-written CUDA using the Warp Matrix Multiply-Accumulate API ($\sim$123~LOC)
  \item \textbf{Raw SIMT} --- Hand-written CUDA without Tensor Cores ($\sim$32~LOC; baseline)
\end{enumerate}

We conduct experiments on three GPUs spanning two architecture generations: NVIDIA H100~NVL (Hopper, \texttt{sm\_90}), RTX~PRO~6000 Blackwell Server Edition (\texttt{sm\_120}), and NVIDIA B200 (Blackwell, \texttt{sm\_100}).
CuTile is supported only on Blackwell GPUs, whereas the other approaches run on all three platforms, providing a controlled setting that isolates CuTile's impact from underlying architectural improvements.

\subsection{Summary of Findings}

Our evaluation yields a clear but nuanced set of conclusions:
\begin{itemize}[nosep]
  \item \textit{Switch for fused attention on datacenter Blackwell (B200).}
    CuTile achieves up to 1{,}007~TFLOP/s---2.5$\times$ higher than FlashAttention-2---using 60~lines of Python code, representing the strongest single result observed in this study.
  \item \textit{Prefer CuTile over WMMA for GEMM on Blackwell GPUs.}
    CuTile delivers 1.5--5.0$\times$ higher throughput than WMMA while requiring 5.6$\times$ less code (22 vs.\ 123~lines), making it a compelling replacement for hand-written WMMA kernels.
  \item \textit{Do not replace cuBLAS for standard GEMM.}
    CuTile achieves 52--79\% of cuBLAS throughput; for workloads already served by \texttt{torch.matmul} or cuBLAS, there is little performance incentive to switch.
  \item \textit{Prefer Triton when cross-architecture portability is required.}
    Triton sustains 62--101\% of cuBLAS performance across all tested GPUs, including Hopper, without architecture-specific tuning.
  \item \textit{Exercise caution on workstation-class Blackwell (\texttt{sm\_120}).}
    On the RTX~PRO~6000, CuTile fused attention reaches only 53\% of FlashAttention-2 throughput, revealing substantial compiler immaturity relative to datacenter Blackwell.
\end{itemize}

% =============================================================================
% 2. BACKGROUND
% =============================================================================
\section{Background: CuTile and Its Alternatives}

\subsection{GPU Architectures Under Evaluation}

Table~\ref{tab:hardware} summarizes the three GPU platforms evaluated in this study.
The H100~NVL (Hopper, \texttt{sm\_90})~\cite{nvidia_h100} represents the current datacenter standard, featuring 132 streaming multiprocessors (SMs) and WGMMA-class Tensor Cores.
The B200 (\texttt{sm\_100})~\cite{nvidia_blackwell} is NVIDIA's next-generation datacenter GPU with 148~SMs and the new Blackwell Tensor Core architecture.
The RTX~PRO~6000 Blackwell Server Edition (\texttt{sm\_120}) is a professional workstation GPU with 188~SMs and a distinct Blackwell variant that differs from B200 in both memory hierarchy and shared-memory configuration.

\begin{table}[t]
\centering
\caption{Hardware platforms used in this study.}
\label{tab:hardware}
\resizebox{\columnwidth}{!}{%
\begin{tabular}{lccc}
\toprule
                        & \textbf{H100 NVL}    & \textbf{RTX PRO 6000}    & \textbf{B200} \\
\midrule
Architecture            & Hopper               & Blackwell                & Blackwell \\
Compute Capability      & \texttt{sm\_90}      & \texttt{sm\_120}         & \texttt{sm\_100} \\
Streaming Multiprocessors & 132                & 188                      & 148 \\
VRAM (GB)               & 94                   & 96                       & 179 \\
TDP (W)                 & 400                  & 600                      & 1{,}000 \\
Max Shared Mem/Block (KB)& 228                 & 48                       & 227 \\
SM Clock (MHz)           & --                   & 2{,}430                  & 1{,}965 \\
CuTile Support          & \textcolor{red}{No}  & \textcolor{green!60!black}{Yes} & \textcolor{green!60!black}{Yes} \\
\bottomrule
\end{tabular}%
}
\end{table}

\subsection{CuTile's Capabilities}

CUDA Tile (CuTile)~\cite{nvidia2025cutile} introduces a tile-centric abstraction for GPU kernel programming via the \texttt{cuda.tile} Python package.
Rather than exposing warps, registers, and shared memory explicitly, CuTile allows developers to express computation in terms of logical tiles while still targeting Tensor Cores and the Tensor Memory Accelerator (TMA) on supported GPUs.
Its core primitives include:
\begin{itemize}[nosep]
  \item \texttt{ct.load} / \texttt{ct.store} --- TMA-backed memory operations that abstract global-to-shared memory transfers
  \item \texttt{ct.mma} --- Tensor Core matrix multiply--accumulate on logical tiles
  \item \texttt{@ct.kernel} with \texttt{ByTarget} --- a mechanism for architecture-specific tuning (e.g., CTA clustering)
\end{itemize}

CuTile's fused attention kernel implements the online softmax algorithm~\cite{milakov2018softmax}, which also underpins FlashAttention-2.
A key constraint for adoption is that CuTile requires a Blackwell GPU (compute capability $\geq$~10.0) and the \texttt{tileiras} compiler introduced in CUDA Toolkit~13.1.
CuTile does not support Hopper (\texttt{sm\_90}) or earlier architectures, in contrast to Triton, which targets a broader range of recent NVIDIA GPUs.
As a result, CuTile cannot currently serve as a single, portable kernel development framework for heterogeneous GPU fleets.

\subsection{Alternatives to CuTile}

\begin{figure*}[t]
\centering
\includegraphics[width=\textwidth]{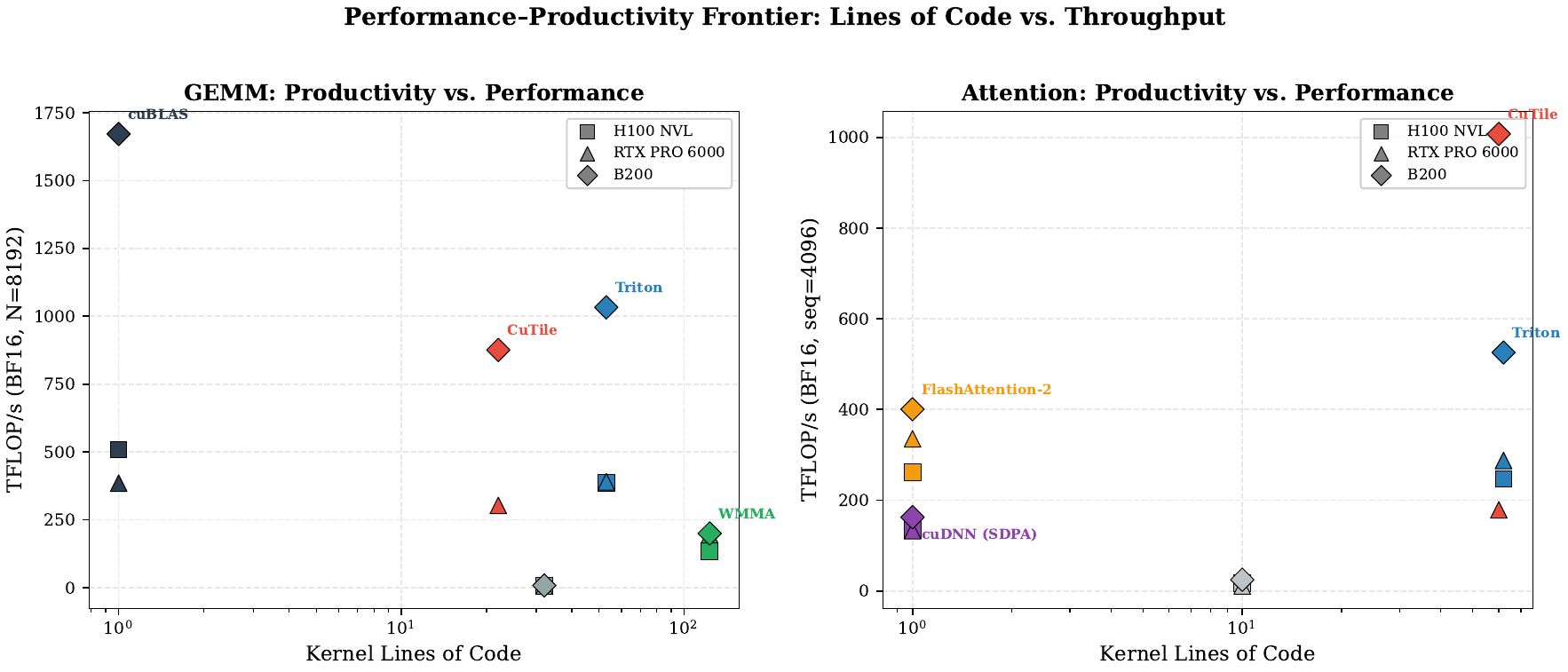}
\caption{Performance--productivity frontier for GEMM (left, $N{=}8192$) and attention (right, seq=4096). Each point represents one abstraction on one GPU. CuTile (red diamonds) offers high productivity but exhibits GPU-dependent performance, while Triton (blue triangles) provides the most consistent portability--performance balance.}
\label{fig:productivity}
\end{figure*}

Figure~\ref{fig:productivity} situates CuTile within the broader performance--productivity design space for both GEMM and fused attention.
For developers considering CuTile, the central question is whether it occupies a region of this design space that is not already covered by existing approaches, including cuBLAS, Triton, WMMA, cuDNN, and FlashAttention-2.

% =============================================================================
% 3. RELATED WORK
% =============================================================================
\section{Related Work: What CuTile Aims to Replace}

\subsection{Hand-Optimized GPU Kernels (Status Quo)}

FlashAttention~\cite{dao2022flashattention} and FlashAttention-2~\cite{dao2023flashattention2} introduced tiled, IO-aware exact attention that avoids materializing the $O(N^2)$ attention matrix, achieving near-peak Tensor Core utilization through carefully hand-optimized CUDA kernels.
Flash-Decoding~\cite{flashdecoding2023} extended these techniques to the autoregressive decode phase.
While these kernels deliver state-of-the-art performance, they consist of thousands of lines of architecture-specific CUDA code and require substantial retuning across GPU generations—precisely the engineering burden CuTile seeks to reduce.

For GEMM, CUTLASS~\cite{cutlass} provides highly optimized C++ templates that form the backbone of production inference systems such as vLLM~\cite{vllm} and TensorRT-LLM~\cite{tensorrt_llm}.
CUTLASS~3.x added support for Hopper features including TMA and WGMMA, but effective use still requires developers to manually specify warp-level tile shapes, shared-memory layouts, and epilogue fusions.
CuTile's \texttt{ct.mma} and \texttt{ct.load}/\texttt{ct.store} primitives explicitly target this complexity by abstracting warp mapping and memory movement while retaining access to Tensor Cores and TMA.

\subsection{Higher-Level Programming Models}

Triton~\cite{tillet2019triton} is the most closely related alternative to CuTile.
It provides a Python-based DSL whose kernels are JIT-compiled via an MLIR-based compiler, with the \texttt{tl.dot} primitive mapping automatically to Tensor Core instructions.
Prior work shows that Triton kernels typically trail hand-optimized CUTLASS by 5--15\% on memory-bound workloads~\cite{vllm}, but Triton supports all recent NVIDIA architectures, making it significantly more portable than CuTile.

Schedule-based frameworks such as TVM and TensorIR~\cite{tvm,tensorir} generate optimized kernels from high-level tensor expressions, while systems like Roller~\cite{roller} and AKG~\cite{akg} further automate the search over implementation spaces.
Halide~\cite{ragan2013halide} pioneered the separation of algorithm and schedule, influencing many of these designs.
Despite strong research results, these systems have not displaced hand-optimized CUDA, CUTLASS, or FlashAttention in production inference engines.

NVIDIA's cuDNN~\cite{chetlur2014cudnn} offers vendor-optimized deep learning primitives, including fused attention, but remains closed-source and does not permit kernel-level customization.
At a lower abstraction level, the WMMA API~\cite{nvidia_wmma} exposes Tensor Cores directly at the warp level, requiring explicit management of threads, registers, and shared memory.
CuTile differs from both approaches by attempting to fully abstract warp-level programming and memory tiling while remaining an officially supported NVIDIA programming model, effectively positioning it as NVIDIA's response to Triton.

\subsection{Cross-Architecture Performance Studies}

GPU kernel benchmarking has a long history, from early dense linear algebra studies~\cite{volkov2008benchmarking} to modern architecture-specific tuning guides~\cite{hopper_tuning_guide,ampere_tuning_guide}.
These studies consistently show that hand-tuned kernels suffer 20--50\% performance degradation when moved across architectures without retuning.

To our knowledge, no prior work has evaluated CuTile's performance portability across GPU architectures.
This paper provides the first empirical study of CuTile across two Blackwell variants (\texttt{sm\_100} and \texttt{sm\_120}) and benchmarks it against abstractions, such as cuBLAS and Triton, that span both Hopper and Blackwell generations.

% =============================================================================
% 4. METHODOLOGY
% =============================================================================

\section{Methodology}

\subsection{Workloads and Microbenchmarks}

We evaluate three kernel families that dominate transformer training and inference: GEMM, fused multi-head attention (FMHA), and end-to-end LLM inference.
All experiments use BF16 or FP16 precision and identical mathematical formulations across implementations.

\paragraph{GEMM.}
We benchmark dense matrix multiplication with shapes representative of both attention and feed-forward network (FFN) layers in transformer models:
\begin{itemize}
  \item Square:
$M = N = K \in \{4096, 8192, 12288, 16384\}$
  \item Rectangular (LLaMA-7B FFN):
$(M, K, N) \in \{2048, 4096\} \times \{(4096, 11008), (11008, 4096)\}$
\end{itemize}

We compare 5 implementations spanning the performance productivity spectrum:
\begin{itemize}[nosep]
  \item \textbf{cuBLAS} \texttt{torch.matmul} (vendor-optimized baseline)
  \item \textbf{Triton} with autotuning over 18 configurations
  \item \textbf{CuTile} using \texttt{ct.mma}
  \item \textbf{WMMA} using \texttt{nvcuda::wmma}, double-buffered \texttt{cp.async}, \texttt{NUM\_STAGES=2}
  \item \textbf{Raw SIMT} using scalar FMA instructions without Tensor Cores
\end{itemize}

\paragraph{Fused Multi-Head Attention (FMHA).}
We evaluate scaled dot-product attention with causal masking under the standard multi-head attention configuration of LLaMA-7B (32 heads, $d{=}128$).
Although many modern systems use grouped-query or multi-query attention for decode efficiency, we focus on standard MHA to isolate compute-bound behavior during prefill.

Experiments use batch size 8 and sequence lengths $\{512, 1024, 2048, 4096, 8192\}$.
Implementations include:
\begin{itemize}[nosep]
  \item FlashAttention-2 (\texttt{flash\_attn}~2.8.3)
  \item cuDNN SDPA (invoked via PyTorch)
  \item Triton (FlashAttention-v2-style kernel with autotuning)
  \item CuTile FMHA (online softmax using \texttt{ct.mma})
  \item Naive unfused attention (three separate GEMMs with softmax)
\end{itemize}

\paragraph{End-to-End LLM Inference.}
To assess system-level impact, we benchmark a LLaMA-7B-like model consisting of 4 transformer layers
($d_{\text{model}}{=}4096$, 32 heads, $d_{\text{head}}{=}128$, $d_{\text{FFN}}{=}11008$) in BF16.
We measure both prefill (batch${\times}$sequence$=$2048) and single-token decode with context length 2048.
Batch sizes are $\{1, 8, 32\}$.

We evaluate four execution modes:
\begin{itemize}[nosep]
  \item Eager execution with naive attention
  \item Eager execution with cuDNN SDPA
  \item Eager execution with FlashAttention-2
  \item \texttt{torch.compile} with SDPA
\end{itemize}

\subsection{Measurement Protocol}

All timings use CUDA event-based measurement
(\texttt{torch.cuda.Event(enable\_timing=True)}), ensuring GPU-side accuracy and excluding host overhead.
Each benchmark runs 10 warmup iterations followed by 50 timed iterations.
Due to its extreme slowness, Raw SIMT uses 3 warmup and 3 timed iterations.

Reported runtimes are arithmetic means of the timed iterations.
We do not report standard deviations, as observed variance was consistently below 1\% across all implementations except Raw SIMT.

Throughput is reported in TFLOP/s, computed as:
\[
\text{GEMM: } \frac{2MNK}{t \cdot 10^{-3} \cdot 10^{12}}, \quad
\text{Attention: } \frac{4BHN^2d}{t \cdot 10^{-3} \cdot 10^{12}}
\]
with the attention FLOP count halved for causal masking.

To ensure correctness, all implementations were validated against reference outputs (e.g., PyTorch and cuBLAS) for numerical equivalence. We additionally performed consistency checks across multiple runs and verified stability of results across different input sizes. Kernel implementations were cross-checked against official documentation and established baselines to minimize the likelihood of implementation errors.

\subsection{Software Environment}

Table~\ref{tab:software} summarizes the software stack used on each platform.
The H100 system runs PyTorch~2.7.1 with CUDA~12.6, while both Blackwell systems use PyTorch~2.8.0 with CUDA~12.8.
CuTile~1.1.0 requires the \texttt{tileiras} compiler provided by CUDA Toolkit~13.1.

\begin{table}[t]
\centering
\caption{Software environment across platforms.}
\label{tab:software}
\resizebox{\columnwidth}{!}{%
\begin{tabular}{lccc}
\toprule
                    & \textbf{H100 NVL} & \textbf{RTX PRO 6000} & \textbf{B200} \\
\midrule
NVIDIA Driver       & 580.105            & 570.195               & 580.126 \\
CUDA Toolkit        & 12.6               & 12.8 + 13.1           & 12.8 + 13.1 \\
PyTorch             & 2.7.1              & 2.8.0                 & 2.8.0 \\
Triton              & 3.3.1              & 3.4.0                 & 3.4.0 \\
FlashAttention      & 2.8.3              & 2.8.3                 & 2.8.3 \\
CuTile              & N/A                & 1.1.0                 & 1.1.0 \\
OS                  & Ubuntu~24.04       & Ubuntu~24.04          & Ubuntu~24.04 \\
\bottomrule
\end{tabular}%
}
\end{table}

% =============================================================================
% 5. GEMM RESULTS
% =============================================================================
\section{GEMM: Can CuTile Replace cuBLAS or Triton?}

\subsection{BF16 Square GEMM}

Table~\ref{tab:gemm-bf16} presents the primary GEMM results across all three GPUs.
The central question for GEMM is whether CuTile's 22-line kernel can approach the performance of cuBLAS (1~line) or Triton (53~lines), which would justify the switch for developers writing custom GEMM variants.

\begin{table*}[t]
\centering
\caption{GEMM performance (TFLOP/s) on BF16 square matrices. CuTile is only available on Blackwell GPUs. Best result per column in \textbf{bold}.}
\label{tab:gemm-bf16}
\begin{tabular}{ll cccc}
\toprule
\textbf{GPU} & \textbf{Implementation} & \textbf{4096$\times$4096} & \textbf{8192$\times$8192} & \textbf{12288$\times$12288} & \textbf{16384$\times$16384} \\
\midrule
\multirow{4}{*}{\rotatebox[origin=c]{0}{\small H100 NVL}}
 & cuBLAS          & \textbf{533.4} & \textbf{508.6} & \textbf{489.5} & \textbf{450.1} \\
 & Triton          & 525.0          & 384.8          & 422.4          & 392.9          \\
 & WMMA            & 113.7          & 132.9          & 121.5          & 118.6          \\
 & Raw SIMT        & 6.3            & 6.2            & 6.2            & 6.2            \\
\midrule
\multirow{5}{*}{\rotatebox[origin=c]{0}{\small RTX PRO 6000}}
 & cuBLAS          & \textbf{378.0} & 384.8          & 386.3          & 386.4          \\
 & Triton          & 346.0          & \textbf{389.9} & \textbf{386.6} & \textbf{386.5} \\
 & CuTile          & 261.1          & 302.8          & 294.5          & 293.9          \\
 & WMMA            & 179.0          & 194.9          & 189.2          & 185.6          \\
 & Raw SIMT        & 5.9            & 5.7            & 5.5            & 5.6            \\
\midrule
\multirow{5}{*}{\rotatebox[origin=c]{0}{\small B200}}
 & cuBLAS          & \textbf{1{,}518.7} & \textbf{1{,}671.8} & \textbf{1{,}557.4} & \textbf{1{,}517.5} \\
 & Triton          & 1{,}066.8      & 1{,}032.9      & 1{,}002.4      & 1{,}011.5      \\
 & CuTile          & 962.2          & 875.8          & 852.0          & 849.9          \\
 & WMMA            & 192.9          & 199.5          & 200.0          & 203.6          \\
 & Raw SIMT        & 8.0            & 7.7            & 7.7            & 7.7            \\
\bottomrule
\end{tabular}
\end{table*}

\begin{figure*}[t]
\centering
\includegraphics[width=\textwidth]{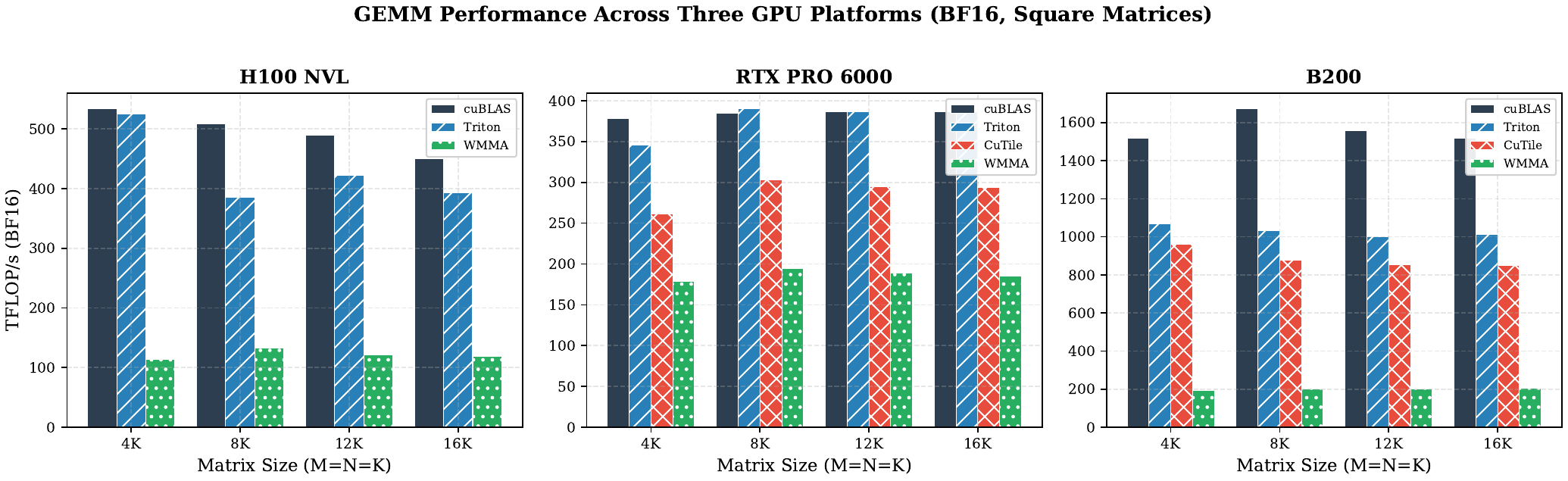}
\caption{GEMM performance (TFLOP/s, BF16) across four square matrix sizes on three GPUs. cuBLAS dominates on all platforms. CuTile (available only on Blackwell) achieves 52--79\% of cuBLAS, significantly outperforming WMMA. Raw SIMT is omitted (6--8 TFLOP/s) for visual clarity.}
\label{fig:gemm-bars}
\end{figure*}

Fig.~\ref{fig:gemm-bars} visualizes these results.

\paragraph{Should you switch to CuTile for GEMM?}
\begin{enumerate}[nosep]
\item \textbf{Not if you're using cuBLAS.}
  cuBLAS dominates on all platforms: 1{,}672~TFLOP/s on B200 (at $N{=}8192$), 533~TFLOP/s on H100, and 385--386~TFLOP/s on RTX~PRO~6000.
  CuTile achieves only 52--79\% of cuBLAS on Blackwell.
  If your workload is a standard GEMM (\texttt{torch.matmul}), there is \emph{no reason to switch}.

\item \textbf{Not if Triton works for your GPU fleet.}
  Triton achieves 98\% of cuBLAS on H100 ($N{=}4096$), 62--70\% on B200, and is \emph{tied} with cuBLAS at larger sizes on RTX~PRO~6000.
  Unlike CuTile, Triton runs on all three GPUs without code changes.

\item \textbf{Yes, if you are currently writing WMMA kernels.}
  CuTile delivers 1.5--5.0$\times$ higher throughput than WMMA on most sizes, in {5.6$\times$ less code} (22 vs.\ 123~lines).
  On the RTX~PRO~6000, CuTile reaches 303~TFLOP/s vs.\ WMMA's 195~TFLOP/s.
  On the B200, CuTile reaches 962~TFLOP/s vs.\ WMMA's 200~TFLOP/s.
  For any developer hand-writing CUDA Tensor Core code, CuTile is a strict upgrade.

\item \textbf{Yes, if you need custom GEMM fusions on Blackwell.}
  CuTile's tile-level API (\texttt{ct.mma} + custom epilogues) makes it practical to write fused GEMM variants that are impossible with cuBLAS and awkward in Triton.
  At 52--79\% of cuBLAS, the performance cost of customizability may be acceptable.

\item \textbf{Raw SIMT is $\sim$64--217$\times$ slower than cuBLAS}, confirming that Tensor Core utilization is essential for competitive GEMM performance and reinforcing the need for abstractions like CuTile.
\end{enumerate}

\subsection{Rectangular GEMM (LLaMA-7B FFN Projections)}

Table~\ref{tab:gemm-rect} shows results for rectangular shapes typical of LLaMA-7B's feed-forward network.

\begin{table}[t]
\centering
\caption{Rectangular GEMM (BF16, TFLOP/s). LLaMA-7B FFN shapes: up = $(M, 4096, 11008)$, down = $(M, 11008, 4096)$.}
\label{tab:gemm-rect}
\resizebox{\columnwidth}{!}{%
\begin{tabular}{ll cc cc}
\toprule
 & & \multicolumn{2}{c}{\textbf{seq=2048}} & \multicolumn{2}{c}{\textbf{seq=4096}} \\
\cmidrule(lr){3-4} \cmidrule(lr){5-6}
\textbf{GPU} & \textbf{Impl.} & \textbf{Up} & \textbf{Down} & \textbf{Up} & \textbf{Down} \\
\midrule
\multirow{2}{*}{H100}    & cuBLAS & 502.6 & 561.2 & 523.4 & 547.3 \\
                          & Triton & 455.2 & 549.1 & 478.3 & 496.9 \\
\midrule
\multirow{3}{*}{RTX PRO} & cuBLAS & 377.4 & 371.1 & 388.8 & 377.4 \\
                          & CuTile & 260.9 & 276.1 & 296.4 & 281.8 \\
                          & Triton & 365.6 & 353.4 & 386.0 & 372.1 \\
\midrule
\multirow{3}{*}{B200}    & cuBLAS & 1{,}448 & 1{,}562 & 1{,}558 & 1{,}562 \\
                          & CuTile & 876.6 & 761.7 & 905.6 & 867.8 \\
                          & Triton & 951.4 & 928.5 & 974.7 & 953.3 \\
\bottomrule
\end{tabular}%
}
\end{table}

The performance ordering is consistent with square GEMM: cuBLAS $>$ Triton $>$ CuTile $>$ WMMA $\gg$ Raw SIMT.
CuTile maintains its position as a strong WMMA replacement in real-world FFN shapes, achieving 261--905~TFLOP/s on Blackwell versus WMMA's 180--204~TFLOP/s.
On the H100 (where CuTile is unavailable), cuBLAS reaches 561~TFLOP/s on the down-projection, which is \emph{higher} than square GEMM due to better L2 cache utilization for the asymmetric shape.

\subsection{FP16 vs.\ BF16 GEMM}

A practical concern for developers considering CuTile is whether precision choice affects the performance gap.
We benchmark FP16 GEMM to answer this.
On the B200, FP16 cuBLAS is within 5\% of BF16 (1{,}597 vs.\ 1{,}672~TFLOP/s at $N{=}8192$), and CuTile FP16 is generally close to BF16 (within 5\% for most sizes, but up to 11\% lower at $N{=}8192$).
This size-dependent sensitivity in the \texttt{tileiras} compiler's FP16 code path is a minor concern: {CuTile's relative standing versus cuBLAS and Triton does not change with precision}.
On the H100, FP16 cuBLAS reaches 498~TFLOP/s at $N{=}4096$ (vs.\ 533~TFLOP/s in BF16, 7\% lower).
On the RTX~PRO~6000, an anomaly emerges: Triton FP16 (378~TFLOP/s) \emph{exceeds} cuBLAS FP16 (310~TFLOP/s) by 22\%, suggesting a cuBLAS tuning gap on \texttt{sm\_120} unrelated to CuTile.

% =============================================================================
% 6. ATTENTION RESULTS
% =============================================================================
\section{Fused Attention: CuTile's Best Case -- and Its Limits}

\subsection{BF16 Causal Attention}

Table~\ref{tab:attn-bf16} presents fused attention results.
If the GEMM story was ``CuTile is good but not the best,'' the attention story is far more dramatic -- and far more GPU-dependent.
This is where CuTile either \emph{dominates} or \emph{disappoints}, depending entirely on which Blackwell chip you have.

\begin{table*}[t]
\centering
\caption{Fused attention performance (TFLOP/s) on BF16 causal attention, batch=8, 32 heads, $d$=128. Best per column in \textbf{bold}. CuTile available only on Blackwell.  --  indicates out-of-memory.}
\label{tab:attn-bf16}
\begin{tabular}{ll cccc c}
\toprule
\textbf{GPU} & \textbf{Implementation} & \textbf{seq=512} & \textbf{seq=1024} & \textbf{seq=2048} & \textbf{seq=4096} & \textbf{seq=8192} \\
\midrule
\multirow{4}{*}{\rotatebox[origin=c]{0}{\small H100 NVL}}
 & FlashAttention-2       & 165.3  & \textbf{222.9}  & \textbf{253.4}  & \textbf{262.0} & \textbf{272.5} \\
 & cuDNN (SDPA)           & 100.8  & 118.4  & 121.8  & 134.2          & 137.2 \\
 & Triton                 & \textbf{168.4}  & 194.1  & 222.0  & 247.2 & 258.8 \\
 & Naive (Unfused)        & 18.8   & 19.3   & 18.2   & 16.6           &  --  \\
\midrule
\multirow{5}{*}{\rotatebox[origin=c]{0}{\small RTX PRO 6000}}
 & FlashAttention-2       & \textbf{177.3}  & \textbf{226.8}  & \textbf{293.2}  & \textbf{335.4} & \textbf{348.6} \\
 & cuDNN (SDPA)           & 89.3   & 106.6  & 125.0  & 132.9          & 137.1 \\
 & Triton                 & 161.3  & 202.2  & 252.5  & 287.7          & 303.3 \\
 & CuTile                 & 87.5   & 127.4  & 157.5  & 178.8          & 191.2 \\
 & Naive (Unfused)        & 9.2    & 10.3   & 10.7   & 10.9           &  --  \\
\midrule
\multirow{5}{*}{\rotatebox[origin=c]{0}{\small B200}}
 & FlashAttention-2       & 207.4  & 297.3  & 362.0  & 400.7          & 422.1 \\
 & cuDNN (SDPA)           & 112.1  & 137.9  & 153.8  & 162.6          & 167.2 \\
 & Triton                 & 234.5  & 362.5  & 461.9  & 525.6          & 562.0 \\
 & \textbf{CuTile}        & \textbf{405.9}  & \textbf{669.9}  & \textbf{887.4}  & \textbf{1{,}007.4} & \textbf{1{,}001.3} \\
 & Naive (Unfused)        & 25.1   & 27.9   & 25.8   & 24.8           &  --  \\
\bottomrule
\end{tabular}
\end{table*}

\begin{figure*}[t]
\centering
\includegraphics[width=\textwidth]{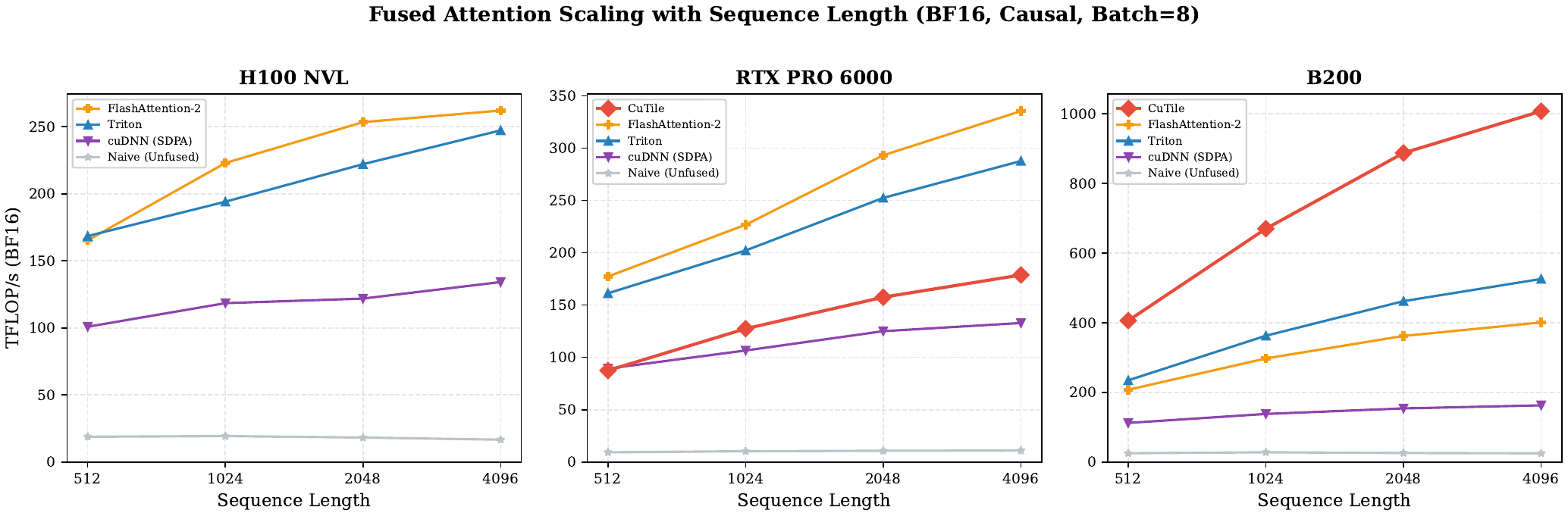}
\caption{Fused attention throughput (TFLOP/s) vs.\ sequence length (BF16, causal, batch=8). On the B200 (right), CuTile (red) dramatically outscales all other implementations, reaching 1{,}007 TFLOP/s at seq=4096. On the RTX PRO 6000 (center), CuTile underperforms FlashAttention-2 and Triton at all sequence lengths.}
\label{fig:attn-scaling}
\end{figure*}

\begin{figure}[t]
\centering
\includegraphics[width=\columnwidth]{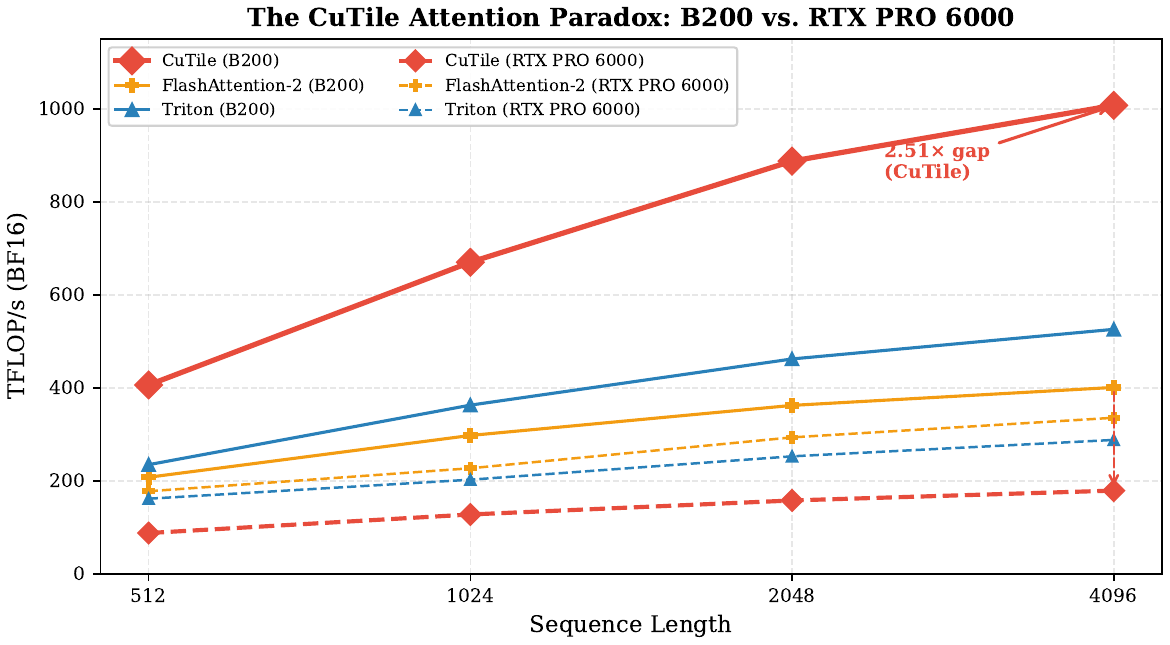}
\caption{The CuTile attention paradox: identical kernel code produces 2.51$\times$ \emph{higher} throughput than FlashAttention-2 on B200 (solid lines) but 47\% \emph{lower} on RTX PRO 6000 (dashed lines). This 5.6$\times$ cross-architecture gap is the largest observed for any abstraction.}
\label{fig:cutile-paradox}
\end{figure}

Fig.~\ref{fig:attn-scaling} shows how each implementation scales with sequence length across the three GPUs, and Fig.~\ref{fig:cutile-paradox} isolates the dramatic CuTile divergence between B200 and RTX~PRO~6000.

\paragraph{CuTile attention is exceptional on B200 but disappointing elsewhere}
The most striking finding of this study -- and the most important for adoption decisions -- is that the \emph{same} CuTile attention kernel produces dramatically different results on two Blackwell GPUs:

\begin{itemize}[nosep]
\item \textbf{B200 (\texttt{sm\_100}) -- Switch to CuTile:} CuTile achieves 1{,}007~TFLOP/s at seq=4096, which is {2.51$\times$ faster} than FlashAttention-2 (401~TFLOP/s) and {1.92$\times$ faster} than Triton (526~TFLOP/s).
This is the highest attention throughput observed in our entire study. For attention-heavy workloads on B200, \emph{CuTile is the best available implementation.}

\item \textbf{RTX~PRO~6000 (\texttt{sm\_120}) -- Do not switch:} CuTile achieves only 179~TFLOP/s at seq=4096, which is {53\% of FlashAttention-2} (335~TFLOP/s) and {62\% of Triton} (288~TFLOP/s). FlashAttention-2 and Triton remain the better choices.
\end{itemize}

We attribute this 5.6$\times$ performance gap between the two Blackwell variants to differences in the \texttt{tileiras} compiler's optimization for \texttt{sm\_100} versus \texttt{sm\_120}, and potentially to differences in the TMA and Tensor Core microarchitectures.
The B200 (\texttt{sm\_100}) appears to be the primary optimization target for the CuTile compiler, consistent with NVIDIA's datacenter-first development strategy.
{This has a critical implication for adoption:} developers must benchmark CuTile on their specific GPU before committing to it.

\paragraph{Without Blackwell, FlashAttention-2 is the safe choice}
FlashAttention-2 provides the most consistent attention performance across all three GPUs, scaling from 165~TFLOP/s (H100, seq=512) to 422~TFLOP/s (B200, seq=8192).
Its hand-optimized CUDA kernels deliver reliable performance regardless of architecture.

\paragraph{Triton performs well on B200}
At seq=4096 on the B200, Triton reaches 526~TFLOP/s, exceeding FlashAttention-2 (401~TFLOP/s) by 1.31$\times$.
For developers who need both attention \emph{and} GEMM performance from a single framework, Triton's consistently strong showing across both workloads makes it a compelling alternative to CuTile.

\subsection{FP16 vs.\ BF16 Attention}

CuTile's relative advantage on B200 persists in FP16: 926~TFLOP/s at seq=4096 (vs.\ 1{,}007 in BF16, 8\% lower), still 2.31$\times$ faster than FlashAttention-2 (400~TFLOP/s in FP16, essentially unchanged from BF16).
The CuTile attention recommendation is precision-invariant: switch on B200, don't switch on RTX~PRO~6000.

% =============================================================================
% 7. END-TO-END INFERENCE
% =============================================================================
\section{End-to-End LLM Inference}

CuTile does not yet provide end-to-end model inference integration -- it offers individual kernel primitives (GEMM, attention) but not a complete inference pipeline. We include these end-to-end results to show what performance levels established PyTorch backends achieve, providing context for when CuTile might become relevant for full-model serving. Table~\ref{tab:e2e} presents end-to-end inference throughput for a LLaMA-7B--like model~\cite{llama} across all three GPUs using standard PyTorch backends. Efficient LLM inference requires careful optimization of both the compute-bound prefill and memory-bound decode phases~\cite{pope2023efficiently}; we focus on the kernel-level performance while noting that advanced techniques such as speculative decoding~\cite{leviathan2023speculative} and continuous batching~\cite{vllm} further improve throughput in production.

\begin{table*}[t]
\centering
\caption{End-to-end LLM inference (LLaMA-7B, 4 layers, BF16). Prefill: TFLOP/s (higher is better). Decode: tokens/sec (higher is better).}
\label{tab:e2e}
\begin{tabular}{ll cc cc cc}
\toprule
 & & \multicolumn{2}{c}{\textbf{H100 NVL}} & \multicolumn{2}{c}{\textbf{RTX PRO 6000}} & \multicolumn{2}{c}{\textbf{B200}} \\
\cmidrule(lr){3-4} \cmidrule(lr){5-6} \cmidrule(lr){7-8}
\textbf{Batch} & \textbf{Backend} & \textbf{Prefill} & \textbf{Decode} & \textbf{Prefill} & \textbf{Decode} & \textbf{Prefill} & \textbf{Decode} \\
 & & (TF/s) & (tok/s) & (TF/s) & (tok/s) & (TF/s) & (tok/s) \\
\midrule
\multirow{4}{*}{1}
 & Eager (Naive)       & 210.8  & 589   & 140.2  & 555   & 374.7  & 767  \\
 & Eager (SDPA)        & 363.6  & \textbf{667}   & 282.8  & 574   & 826.9  & 866  \\
 & Eager (FA2)         & 364.0  & 527   & 283.3  & \textbf{575}   & \textbf{837.3}  & 865  \\
 & torch.compile       & \textbf{364.2}  & 630   & \textbf{283.5}  & 574   & 827.5  & \textbf{866}  \\
\midrule
\multirow{4}{*}{8}
 & Eager (Naive)       & 204.1  & 1{,}215  & 139.7  & 1{,}206  & 389.2  & 1{,}582  \\
 & Eager (SDPA)        & 361.1  & \textbf{2{,}137}  & 303.6  & 2{,}123  & 910.0  & 3{,}113  \\
 & Eager (FA2)         & \textbf{367.9}  & 2{,}133  & \textbf{303.8}  & \textbf{2{,}124}  & \textbf{918.0}  & \textbf{3{,}142}  \\
 & torch.compile       & 357.7  & 2{,}123  & 303.2  & 2{,}124  & 905.9  & 3{,}083  \\
\midrule
\multirow{4}{*}{32}
 & Eager (Naive)       & 197.4  & 1{,}433  & 141.5  & 1{,}482  & 381.6  & 1{,}924  \\
 & Eager (SDPA)        & \textbf{350.7}  & 2{,}851  & 303.4  & 3{,}029  & 920.5  & 4{,}601  \\
 & Eager (FA2)         & 349.7  & \textbf{2{,}887}  & \textbf{303.4}  & \textbf{3{,}030}  & \textbf{924.0}  & \textbf{4{,}647}  \\
 & torch.compile       & 346.0  & 2{,}840  & 302.4  & 3{,}029  & 920.4  & 4{,}594  \\
\bottomrule
\end{tabular}
\end{table*}

Below are the key findings and CuTile implications.

\paragraph{Fused attention backends (SDPA, FA2) provide 1.7--2.4$\times$ prefill speedup over the naive unfused baseline across all GPUs} Given CuTile's 2.51$\times$ attention advantage on B200, a CuTile-integrated inference pipeline could potentially further improve prefill throughput beyond what SDPA/FA2 achieve.

\paragraph{torch.compile provides minimal additional benefit for this 4-layer model configuration} At batch=1 on the H100, torch.compile achieves 364.2~TFLOP/s versus 364.0 for Eager~(FA2) -- a negligible difference. This suggests that kernel-level optimization (where CuTile operates) matters more than graph-level compilation for well-optimized models.

\paragraph{Decode performance is memory-bandwidth bound and remarkably uniform across attention backends once fused attention is used} On the B200 at batch=32, all three fused backends achieve 4{,}594--4{,}647 tok/s. CuTile would provide no decode advantage here, as decode is bottlenecked by memory bandwidth, not compute.

\paragraph{B200 dominates in absolute throughput} 924~TFLOP/s prefill and 4{,}647~tok/s decode at batch=32, compared to 350~TFLOP/s and 2{,}887~tok/s on the H100. This 2.6$\times$ prefill advantage reflects the B200's higher SM count and memory bandwidth -- the same platform where CuTile attention excels.

% =============================================================================
% 8. CODE COMPLEXITY
% =============================================================================
\section{The Productivity Payoff: How Much Code Does CuTile Save?}

Performance is only half the adoption decision.
The other half is developer productivity -- how much engineering effort does each abstraction demand?
Table~\ref{tab:loc} compares lines of code across all implementations.

\begin{table}[ht!]
\centering
\caption{Lines of code comparison. Kernel LOC = GPU kernel only; Total LOC includes launcher/wrapper code.}
\label{tab:loc}
\resizebox{\columnwidth}{!}{%
\begin{tabular}{lll rr}
\toprule
\textbf{Operation} & \textbf{Implementation} & \textbf{Abstraction} & \textbf{Kernel} & \textbf{Total} \\
\midrule
\multirow{5}{*}{GEMM}
 & cuBLAS (\texttt{torch.matmul})   & Library API     & 1   & 1 \\
 & Triton                           & Python DSL      & 53  & 76 \\
 & CuTile                           & Python DSL      & 22  & 45 \\
 & WMMA (CUDA C++)                  & CUDA C++        & 123 & 183 \\
 & Raw SIMT (CUDA C++)              & CUDA C++        & 32  & 58 \\
\midrule
\multirow{5}{*}{Attention}
 & FlashAttention-2                 & Library API     & 1   & 1 \\
 & PyTorch SDPA                     & Library API     & 1   & 4 \\
 & Triton                           & Python DSL      & 62  & 87 \\
 & CuTile FMHA                      & Python DSL      & 60  & 95 \\
 & Naive (Unfused)                  & PyTorch eager   & 10  & 10 \\
\bottomrule
\end{tabular}%
}
\end{table}

\paragraph{CuTile's strongest argument: the GEMM kernel is 22~lines}
At 22~lines for the GEMM kernel, CuTile requires {5.6$\times$ less code than WMMA} (123 lines) and {2.4$\times$ less than Triton} (53 lines).
CuTile's \texttt{ct.load}, \texttt{ct.mma}, and \texttt{ct.store} primitives abstract the entire shared memory tiling, register allocation, and warp scheduling pipeline that must be manually managed in WMMA.
For developers currently maintaining WMMA kernels, this productivity gain alone may justify the switch.

\paragraph{For attention, CuTile offers no productivity advantage over Triton}
Both require comparable effort (60 vs.\ 62 kernel LOC), as both implement the same Flash Attention v2 algorithm with online softmax.
The similar code sizes reflect that attention's algorithmic complexity (causal masking, running max/sum for online softmax) dominates over the tiling abstractions.
The decision between CuTile and Triton for attention therefore comes down purely to performance (CuTile wins on B200) and portability (Triton wins everywhere else).

\paragraph{The adoption tradeoff}
CuTile occupies a unique position: it offers near-library-call brevity while exposing enough control for kernel customization.
For teams targeting \emph{only Blackwell datacenter GPUs}, CuTile provides the best productivity-to-performance ratio.
For teams with mixed GPU fleets (Hopper + Blackwell, or datacenter + workstation), Triton provides better portability despite requiring up to 2.4$\times$ more kernel code for GEMM (53 vs.\ 22 lines; attention code sizes are comparable at 62 vs.\ 60 lines).

% =============================================================================
% 9. ANALYSIS & DISCUSSION
% =============================================================================
\section{Decision Framework: When to Use CuTile}

\subsection{Understanding CuTile's GPU Sensitivity}

The 5.6$\times$ performance gap between CuTile attention on B200 (1{,}007~TFLOP/s) and RTX~PRO~6000 (179~TFLOP/s) is the most important finding for adoption decisions.
Before switching to CuTile, developers must understand \emph{why} this gap exists:

\paragraph{Compiler maturity} The \texttt{tileiras} compiler (CUDA Toolkit 13.1) is optimized primarily for \texttt{sm\_100} (datacenter Blackwell). The \texttt{sm\_120} (workstation Blackwell) appears to receive less compiler optimization.

\paragraph{Micro-architectural differences} Despite both being ``Blackwell,'' \texttt{sm\_100} and \texttt{sm\_120} have significant differences. The RTX~PRO~6000 has only 48~KB shared memory per block (vs.\ 227--228~KB opt-in on B200/H100), which constrains tile sizes and double-buffering strategies that CuTile relies on.

\paragraph{CTA clustering sensitivity} CuTile uses CTA (Cooperative Thread Array) clustering via the \texttt{ByTarget} decorator. Our experiments showed that the optimal CTA cluster size differs between \texttt{sm\_100} and \texttt{sm\_120}, and incorrect configuration can cause 4$\times$ slowdowns.

The practical implication is that CuTile should be considered mature for \texttt{sm\_100} (B100, B200) but \emph{experimental} for \texttt{sm\_120} (RTX PRO series).

\subsection{The Normalized Picture}

\begin{figure*}[ht!]
\centering
\includegraphics[width=\textwidth]{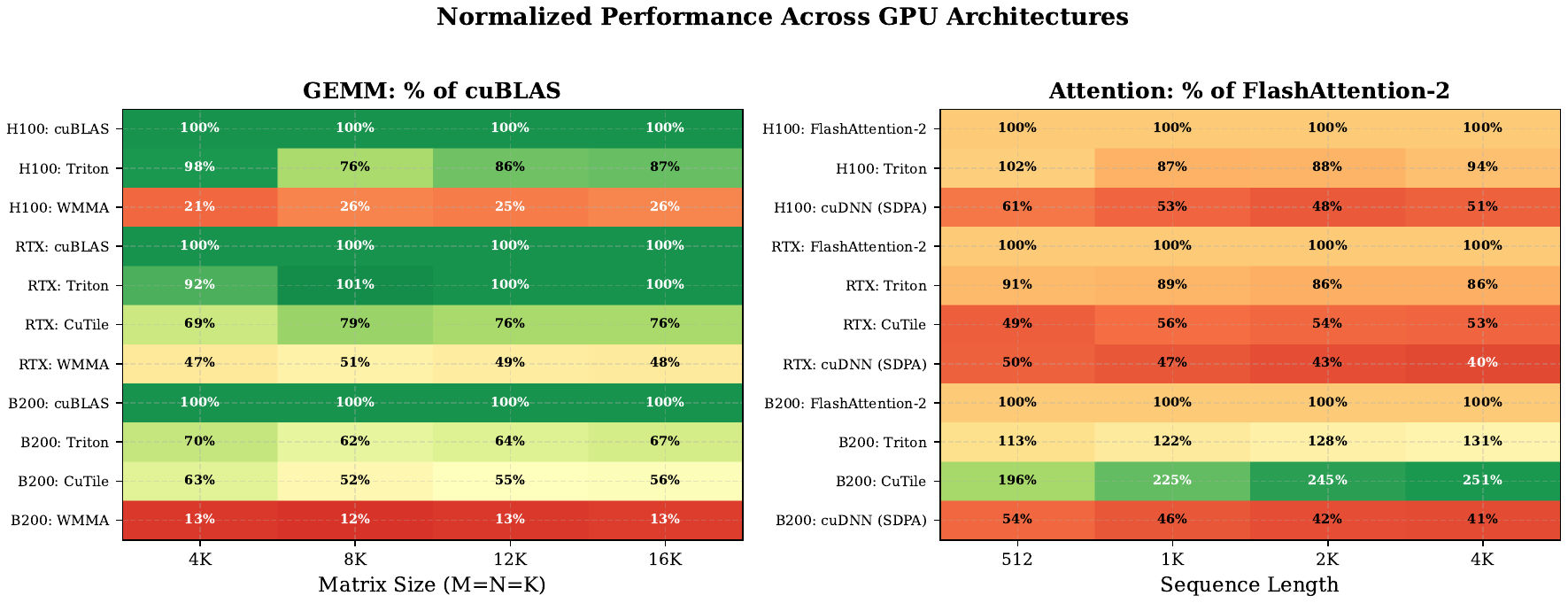}
\caption{Normalized performance heatmap. Left: GEMM throughput as percentage of cuBLAS. Right: attention throughput as percentage of FlashAttention-2. CuTile on B200 achieves up to 251\% of FA2 (dark green), while CuTile on RTX PRO 6000 reaches only 53\% (orange-red). This GPU-dependent behavior is unique to CuTile among all tested abstractions.}
\label{fig:normalized}
\end{figure*}

Fig.~\ref{fig:normalized} presents a normalized performance heatmap across all GPU--implementation combinations, making the adoption decision visual.

For GEMM, the hierarchy is consistent: cuBLAS $>$ Triton $>$ CuTile $>$ WMMA $\gg$ Raw SIMT on every GPU.
This means the GEMM adoption decision is stable regardless of target hardware.
For attention, the hierarchy \emph{changes between GPUs} -- CuTile is best on B200 but worst-among-fused on RTX~PRO~6000 -- making the adoption decision hardware-dependent.

\subsection{CuTile vs.\ Each Alternative: A Direct Comparison}

\begin{table}[ht!]
\centering
\caption{When to choose CuTile over each alternative.}
\label{tab:portability}
\resizebox{\columnwidth}{!}{%
\begin{tabular}{lp{6.5cm}}
\toprule
\textbf{Instead of \ldots} & \textbf{Switch to CuTile when \ldots} \\
\midrule
cuBLAS  & You need custom fusions that cuBLAS doesn't provide (e.g.  fused GEMM+activation) \\
Triton  & You target only B200/B100 \emph{and} need attention kernels (2.5$\times$ advantage) \\
WMMA    & Always, on any Blackwell GPU (1.5--5$\times$ faster, 5.6$\times$ less code) \\
FA2     & Only on B200/B100 for attention; FA2 is better on RTX PRO and all Hopper \\
Raw SIMT & Always (64--217$\times$ faster with Tensor Cores) \\
\bottomrule
\end{tabular}%
}
\end{table}

\begin{figure}[t]
\centering
\includegraphics[width=\columnwidth]{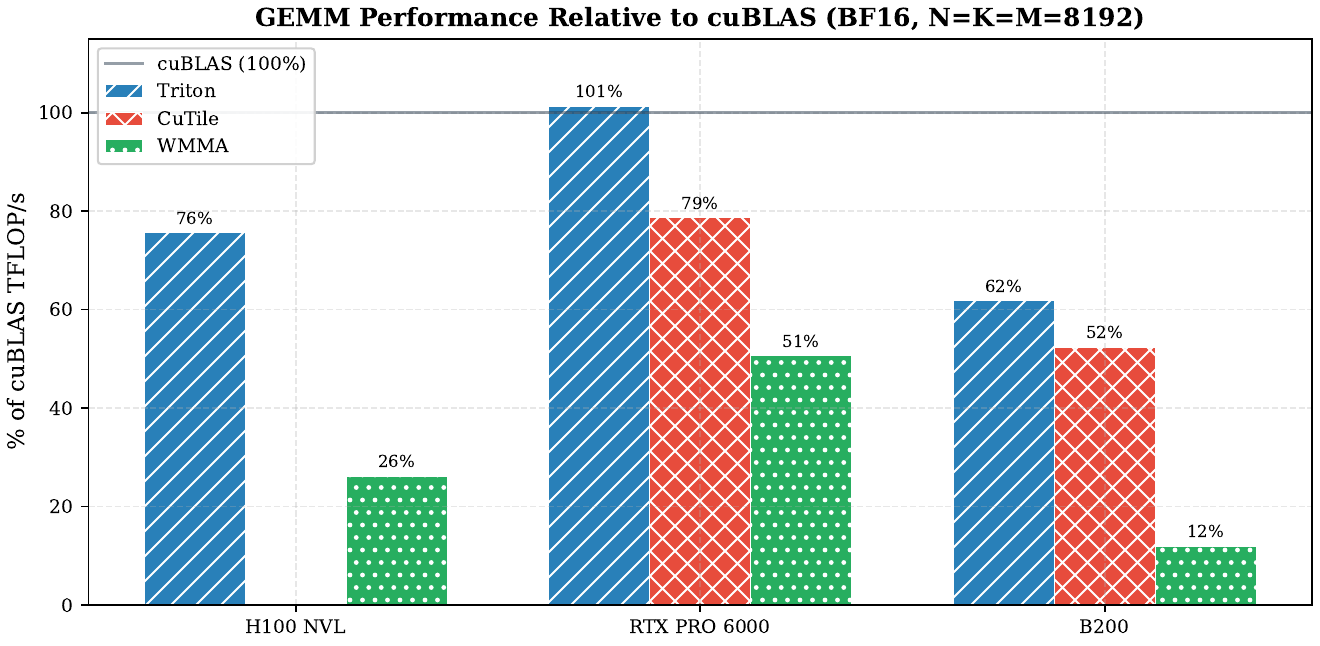}
\caption{GEMM performance as a percentage of cuBLAS at $N{=}8192$ across three GPUs. Triton (blue) achieves the most consistent cross-architecture performance (62--101\%). CuTile (red) shows 52\% on B200 and 79\% on RTX PRO, but is unavailable on H100.}
\label{fig:cross-gpu}
\end{figure}

Table~\ref{tab:portability} summarizes when CuTile is the better choice over each alternative.
Fig.~\ref{fig:cross-gpu} visualizes the relative GEMM performance, showing CuTile's position in the ecosystem.

\subsection{CuTile's Scaling Advantage for Long Sequences}

One area where CuTile shows uniquely strong behavior is scaling with problem size.
On the B200, CuTile attention throughput \emph{increases} from 406~TFLOP/s (seq=512) to 1{,}007~TFLOP/s (seq=4096), a 2.48$\times$ improvement that suggests excellent pipeline utilization at larger tile counts.
In contrast, FlashAttention-2 shows more modest scaling (207$\to$401~TFLOP/s, 1.93$\times$).
{For long-context LLM applications on B200}, CuTile's advantage grows with sequence length, making it increasingly attractive as context windows expand.

\subsection{Limitations and Caveats}

Developers considering CuTile should be aware of these limitations in our evaluation:

\paragraph{Single-GPU evaluation} We benchmark 1 instance per GPU.
  Manufacturing variation could affect results by 1--3\%.
  
\paragraph{No Nsight Compute profiling} We report application-level TFLOP/s but do not decompose into Tensor Core utilization or memory bandwidth. Future work should include roofline analysis~\cite{williams2009roofline}.

\paragraph{CuTile compiler is evolving} The \texttt{tileiras} compiler from CUDA Toolkit 13.1 is still maturing.
  Future versions may significantly improve \texttt{sm\_120} performance, potentially changing our RTX~PRO~6000 recommendation.
  
\paragraph{No FP8/INT8} All experiments use BF16/FP16.
  CuTile's advantage for quantized inference (increasingly important for production) is unknown.
  
\paragraph{4-layer proxy model} Our end-to-end inference uses a 4-layer proxy, not a full 32-layer LLaMA-7B.
  A CuTile-integrated full model may exhibit different bottlenecks.
  
\paragraph{Blackwell-only} CuTile cannot run on Hopper, Ampere, or earlier GPUs. Teams with mixed fleets need a dual-framework strategy.

\paragraph{Software version differences} The H100 runs PyTorch~2.7.1 with CUDA~12.6, while both Blackwell GPUs run PyTorch~2.8.0 with CUDA~12.8. These differences could introduce confounds in cross-GPU comparisons for Triton and FlashAttention-2, though we expect the effect to be small ($<$2\%) as both versions use the same core algorithms.

Given the rapid evolution of GPU architectures and the CuTile compiler stack, the results presented in this study reflect the current state of the ecosystem and may change with future CUDA releases and hardware generations.

% =============================================================================
% 10. CONCLUSION
% =============================================================================
\section{Conclusion: Should You Switch to CuTile?}

We have evaluated CuTile head-to-head against cuBLAS, Triton, WMMA, and Raw~SIMT across NVIDIA's Hopper (H100) and Blackwell (B200, RTX~PRO~6000) architectures on three workloads: GEMM, fused attention, and end-to-end LLM inference.
Below is our recommendations.

\subsection{Switch to CuTile If}
\paragraph{Writing fused attention kernels for B200/B100 datacenter GPUs} CuTile delivers 1{,}007~TFLOP/s -- 2.5$\times$ faster than FlashAttention-2 -- in 60~lines of Python. No other abstraction comes close.
\paragraph{Currently maintaining WMMA kernels and your deployment targets Blackwell} CuTile provides 1.5--5.0$\times$ better throughput in 5.6$\times$ less code. It is a strict upgrade.
\paragraph{Need custom GEMM fusions (e.g. fused GEMM + activation, custom epilogues) on Blackwell and cannot use cuBLAS's fixed API}
CuTile's 22-line kernel is far more accessible than WMMA's 123 lines.

\subsection{Not to Switch If}
\paragraph{Need {cross-architecture portability} (Hopper + Blackwell, or mixed GPU fleets)} CuTile does not run on Hopper or earlier architectures. Triton is a good option (62--101\% of cuBLAS on all GPUs, no architecture-specific code changes).

\paragraph{Using {standard GEMM} (\texttt{torch.matmul} / cuBLAS) and do not need kernel customization} CuTile achieves only 52--79\% of cuBLAS. There is no reason to switch.

\paragraph{Targeting RTX~PRO~6000 or other workstation Blackwell GPUs (\texttt{sm\_120})} CuTile attention achieves only 53\% of FlashAttention-2 on this platform due to compiler immaturity.

\paragraph{Need {FP8 or INT8} kernels} CuTile's quantized performance is untested.

\subsection{Wait and Re-evaluate When}
\paragraph{CUDA Toolkit 14.x or later is released}
  The \texttt{tileiras} compiler is still maturing. Our RTX~PRO~6000 results may improve dramatically with future compiler versions, potentially resolving the 5.6$\times$ cross-GPU gap.
\paragraph{CuTile gains Hopper support}
  If NVIDIA backports CuTile to \texttt{sm\_90}, the portability objection disappears and CuTile becomes directly competitive with Triton.
\paragraph{CuTile is integrated into PyTorch or vLLM}
  End-to-end inference frameworks that use CuTile attention on B200 could unlock the 2.5$\times$ attention speedup at the system level.

\subsection{The broader lesson}
CuTile is not yet a replacement for established abstractions, but it is the first credible vendor-backed attempt to make Tensor Core programming \emph{accessible} to Python developers.
The fact that a 60-line CuTile kernel outperforms FlashAttention-2's thousands of lines of hand-optimized CUDA on B200 is a paradigm shift.
As the compiler matures and GPU support broadens, tile-centric programming models like CuTile may fundamentally change how GPU kernels are developed.

\vspace{2mm}
\noindent
The artifacts are located at:
\url{https://github.com/uwm-se/CuTile}.

% =============================================================================
% REFERENCES
% =============================================================================
\bibliographystyle{IEEEtran}

\end{document}